\documentclass{article}
\PassOptionsToPackage{numbers,sort&compress}{natbib}

\usepackage[preprint]{neurips_2020}

\usepackage[utf8]{inputenc} 
\usepackage[T1]{fontenc}    
\usepackage{hyperref}       
\usepackage{url}            
\usepackage{booktabs}       %
\usepackage{nicefrac}       
\usepackage{microtype}      
\usepackage{soul} 
\usepackage{amsmath,amsfonts,amssymb,amsbsy}
\usepackage{url}
\usepackage{xcolor}
\usepackage{footnote}
\usepackage{subcaption}
\usepackage{wrapfig}
\usepackage{multirow}
\usepackage{tikz}
\usepackage{placeins}
\usepackage{bm}
\usepackage{comment}
\usepackage{algorithm}
\usepackage[noend]{algpseudocode}
\setcitestyle{square,numbers}
\usepackage[titletoc]{appendix}
\usepackage{todonotes}
\usepackage[capitalize]{cleveref}

\DeclareMathOperator*{\argmin}{arg\,min}

\title{
Robust, Accurate Stochastic Optimization for Variational Inference}

\author{%
  Akash Kumar Dhaka  \\
 \texttt{akash.dhaka@aalto.fi} 
  \And
  Alejandro Catalina  \\
  \texttt{alejandro.catalina@aalto.fi} \\
  \AND
  Michael Riis Andersen \\
  \texttt{miri@dtu.dk} \\
  \And
  Mans Magnusson \\
  \texttt{mans.magnusson@aalto.fi} \\
  \And Jonathan H. Huggins \\
  \texttt{huggins@bu.edu}
  \And
  Aki Vehtari \\
  \texttt{aki.vehtari@aalto.fi} \\
}

\begin{document}
\maketitle
\begin{abstract}
We consider the problem of fitting variational posterior approximations using stochastic optimization methods. The performance of these approximations depends on (1) how well the variational family matches the true posterior distribution, (2) the choice of divergence, and (3) the optimization of the variational objective. 
We show that even in the best-case scenario when the exact posterior belongs to the assumed variational family,
common stochastic optimization methods lead to poor variational approximations if the problem dimension is moderately large. 
We also demonstrate that these methods are not robust across diverse model types. 
Motivated by these findings, we develop a more robust and accurate stochastic optimization framework by viewing the underlying optimization algorithm as producing a Markov chain.
Our approach is theoretically motivated and includes a diagnostic for convergence and a novel stopping rule, 
both of which are robust to noisy evaluations of the objective function. 
We show empirically that the proposed framework works well on a diverse set of models: it can automatically detect 
stochastic optimization failure or inaccurate variational approximation.
\end{abstract}

\section{Introduction} \label{sec:intro}

Bayesian inference is a popular approach due to its flexibility and theoretical foundation in probabilistic reasoning \citep{Robert:2007,BS:2000}. 
The central object in Bayesian inference is the posterior distribution of the parameter of interest given the data.
However, using Bayesian methods in practice usually requires approximating the posterior distribution. 
Due to its computational efficiency, variational inference (VI) has become a commonly used approach for large-scale approximate inference in machine learning~\cite{Jordan99anintroduction, wainwright_jordan08}. 
Informally, VI methods find a simpler approximate posterior that minimizes a divergence measure $\mathcal{D}\left[q||p\right]$ 
from the approximate posterior $q$ to the exact posterior distribution $p$ 
-- that is, they compute a optimal variational approximation $q^* = \arg\min_{q \in \mathcal{Q}} \mathcal{D}\left[q||p\right]$.
The variational family is often parametrized by a vector $\bm{\lambda} \in \mathbb{R}^K$ 
so the parameter of $q^*$ is given by
\begin{align}\label{lambdastar}
\bm{\lambda}^* = \argmin_{\bm{\lambda} \in \mathbb{R}^K} \mathcal{D}\left[q_{\bm{\lambda}}||p\right].
\end{align}
Variational approximations in machine learning is typically used for prediction,
but recent work has shown that these approximations possess good statistical properties as point estimators and as  
posterior approximations \cite{Wang:2018,Wang:2019:VB-misspecified,Pati:2018,CheriefAbdellatif:2018}. 
Variational inference is therefore becoming an attractive statistical method since variational approximations 
can often be computed more efficiently than either the maximum likelihood estimate or more precise posterior estimates
-- particularly when there are local latent variables that need to be integrated out. 
Therefore, there is a need to develop variational methods that are appropriate for statistical inference: where the model parameters are 
themselves the object of interest, and thus the accuracy of the approximate posterior compared to the true posterior is important.
In addition, we would ideally like to refine a variational approximation further using importance sampling \cite{yao18a,Huggins:2020:VI} -- as in the adaptive importance sampling literature \citep{Oh:1989}.

\begin{figure}[tp]
\begin{subfigure}[t]{.495\textwidth}
\includegraphics[width=0.99\textwidth]{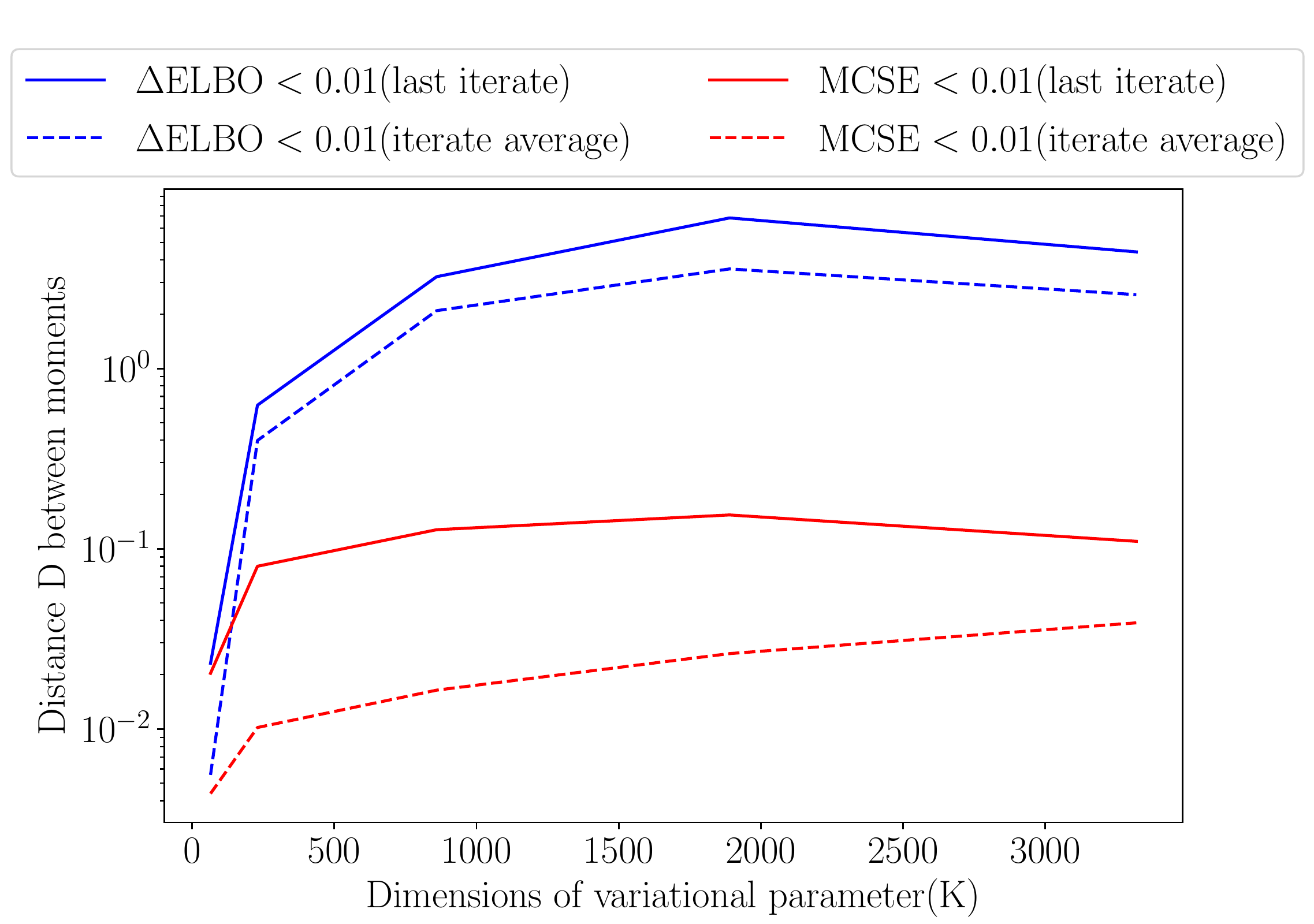}
\label{fig:khatvsD}
\end{subfigure}
\hfill
\begin{subfigure}[t]{.495\textwidth}
\includegraphics[width=0.99\textwidth]{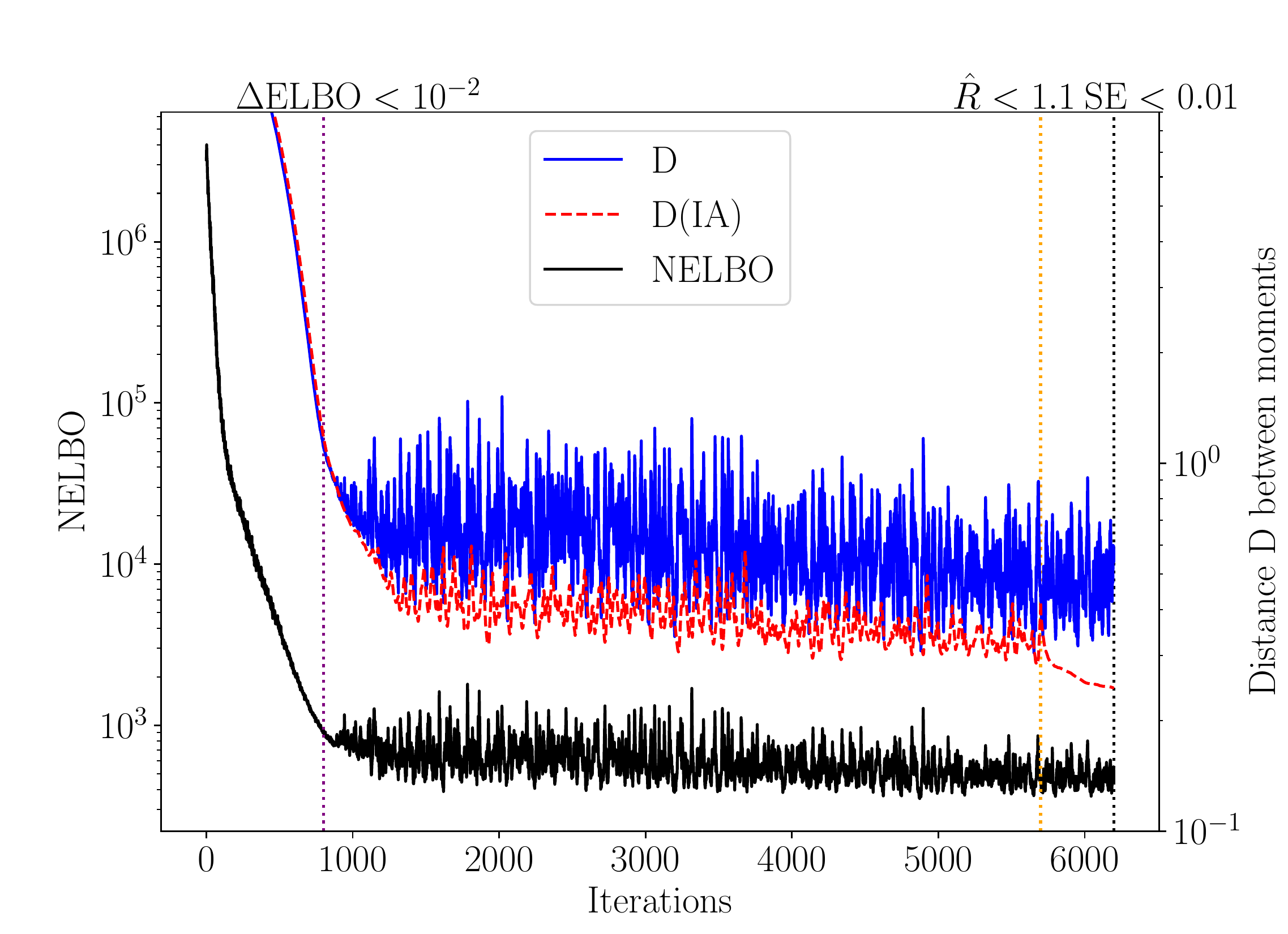}
\label{fig:stopping-rule}
\end{subfigure}
\caption{
{\textbf{(left)} The distance between the variational and ground truth moments 
for a full rank VI approximation on linear regression models of varying dimensions of posterior. 
$\Delta \textrm{ELBO}$ denotes the standard stopping rule, MCSE denotes our proposed stopping rule, and
IA indicates that our iterate averaging approach was used. 
IA and our proposed stopping rule both improve accuracy, particularly in higher dimensions. 
\textbf{(right)} The negative ELBO (NELBO) and the distances between the variational and ground truth moments based 
on the current iterate and using IA.
The stopping point based on $\Delta \textrm{ELBO}$ is shown by the dotted red line
and occurs prematurely. 
Using our proposed algorithm, the starting and stopping points for IA are shown by the dotted orange and black lines, respectively.}}
\label{fig:intro-example}
\end{figure}

Meanwhile, two developments have greatly increased the scope of the applicability of VI methods.
The first is stochastic variational inference (SVI), where  \cref{lambdastar} 
is solved using stochastic optimization with mini-batching~\cite{hoffman13}. 
The increased computational efficiency of mini-batching allows SVI to scale to datasets with tens of millions of observations. 
The second is black box variational inference methods, which have extended variational inference to a wide range of models in probabilistic programming context
by removing the need for model-specific derivations~\cite{ranganath14, Kucukelbir15, titsias14}. 
This flexibility is obtained by approximating local expectations and their auto-differentiated gradients using  Monte Carlo approximations. 
While using stochastic optimization to solve \cref{lambdastar} makes variational inference scalable as well as flexible, 
there is a drawback: it becomes increasingly difficult to solve the optimization problem with sufficiently high accuracy,
particularly as the dimensionality of the variational parameter $\bm{\lambda}$ increases.
\Cref{fig:intro-example}(left, solid lines) demonstrates this phenomenon on a simple linear 
regression problem where the exact posterior belongs to the variational family.
Since $q^* = p$, all of the error is due to the stochastic optimization.

Because in machine learning the quality of a posterior approximation is usually evaluated by out-of-sample predictive performance,
the additional error from the stochastic optimization does not necessarily problematic. 
Therefore, there has been less attention paid to developing stochastic optimization schemes that provide very accurate variational parameter estimates and, ideally, have good importance sampling properties too. 
And, as seen in \cref{fig:intro-example}(left, solid blue line), standard VI optimization schemes remain insufficient for 
statistical inference because they do not provide accurate variational parameter estimates
-- particularly in higher dimensions.


Moreover, existing optimizers are fragile, in that they require the choice of many hyperparameters and can fail badly.
For example, the common stopping rule $\Delta$ELBO \cite{Kucukelbir15} is based on the change in the 
variational objective function value (the negative ELBO).
But, as illustrated in \cref{fig:intro-example}(right), using $\Delta$ELBO results in termination 
before the optimizer converges, resulting in an inaccurate variational approximation (intersection of blue line and purple vertical line). 
Using a smaller cutoff for $\Delta$ELBO to ensure convergence resulted in the criterion never being met
because the stochastic estimates of the negative ELBO were too noisy.
To remedy this problem a combination of a smaller step size (resulting in slower convergence) 
and a more accurate Monte Carlo gradient estimates (resulting is greater per-iteration computation) must be used.
Thus, the standard optimization algorithm is fragile due to a non-trivial interplay between its many hyperparameters, 
which requires the user to carefully tune all of them jointly. 

In this paper, we address the shortcomings of current stochastic optimizers for VI by viewing 
the underlying optimization algorithm as producing a Markov chain.
While such a perspective has been pursued in theoretical contexts \cite{Erdogdu:2018,Raginsky:2017} 
and in the deep neural network literature~\cite{huang17,izmailov18,garipov18,maddox2019simple}, 
the potential innovative algorithmic consequences of such a perspective, particularly in the VI context, have not been explored.
Our Markov chain perspective allows us create more accurate variational parameter estimates by using iterate averaging, which is
particularly effective in high dimensions (see red dotted lines in \cref{fig:intro-example}).
But, even when using iterate averaging, the problems of fragility remain.
In particular, we need to decide (A) when to start averaging (or when the optimizer has failed) and (B) when to terminate the optimization. 
For (A), we use the $\widehat{R}$ diagnostic \cite{gelman92,vehtari2019rhat}, a well-established method from the MCMC literature. 
For (B), we use Monte Carlo standard error estimates based on the chain's effective sample size (ESS) and the ESS itself~\cite{vehtari2019rhat}
to ensure convergence of the parameter estimate (again drawing on a rich MCMC literature~\cite{flegal08a, flegal08b}).
We also use the $\hat{k}$ diagnostic from the importance sampling literature to check on the quality of the variational approximation and determine
whether it can be used as an importance distribution \cite{vehtari2019pareto,yao18a}.
By combining all of these ideas, we develop an optimization framework that
is robust to the selection of optimization hyperparameters such as step size and mini-batch size
while also producing substantially more accurate posterior approximations. 
We  empirically validate our proposed framework on a wide variety of models and datasets. 
\section{Background: Variational Inference} \label{sec:background}
Let $p(\bm{y}, \bm{\theta})$ denote the joint density for a model of interest, where $\bm{y} \in \mathcal{Y}^N$ is a vector of $N$ observations and $\bm{\theta} \in \mathbb{R}^P$ is a vector of model parameters. In this work, we assume that the observations are conditionally independent given $\bm{\theta}$; that is, the joint density factorizes as\footnote{In addition, we may have that $p(y_i| \bm{\theta}) = \int p(y_i | \bm{\theta}, z_i)p(z_i | \bm{\theta})d z_i $. But, for simplicity, we do not write the explicit dependence on the local latent variable $z_i$.}
$p(\bm{y}, \bm{\theta}) = \prod_{i=1}^N p(y_i| \bm{\theta}) p_0(\bm{\theta})$.
The goal is to approximate the resulting posterior distribution, $p(\bm{\theta}) \equiv p(\bm{\theta}|\bm{y})$, 
by finding the best approximating distribution $q \in \mathcal{Q}$ in the variational family $\mathcal{Q}$ as measured by a divergence measure.
We focus on two commonly used variational families -- the \textit{mean-field} and the \textit{full-rank} Gaussian families --
and the standard Kullback--Leibler (KL) divergence objective, but our approach generalizes to other variational families and divergences as well.
It can be shown that minimizing the KL divergence is equivalent to maximizing the functional known as the evidence lower bound (ELBO) $\mathcal{L}: \mathbb{R}^K \rightarrow \mathbb{R}$ given by~\citep{blei17}
\begin{align} 
\mathcal{L}\left(\bm{\lambda}\right) &\equiv \mathbb{E}_{q} \left[\ln p(\bm{y}, \bm{\theta})\right] - \mathbb{E}_q\left[\ln q (\bm{\theta}) \right]
= \sum_{i=1}^N \left(\mathbb{E}_{q} \left[\ln p(y_i| \bm{\theta})\right] - \frac{1}{N}\text{KL}\left[q||p_0\right]\right) = \sum_{i=1}^N \nonumber \mathcal{L}_i\left(\bm{\lambda}\right),
\end{align}
where $q$ is parametrized by $\bm{\lambda} \in \mathbb{R}^K$ and $\mathcal{L}_i(\bm{\lambda}) \equiv \mathbb{E}_{q} \left[\ln p(y_i| \bm{\theta})\right] - \frac{1}{N}\text{KL}\left[q||p_0\right]$. 
The optimal approximation is $q_{\bm{\lambda}^*}$ for $\bm{\lambda}^* = \arg\max_{\bm{\lambda}} \mathcal{L}\left(\bm{\lambda}\right)$. 

\subsection{Stochastic Optimization for VI}


We will consider approximately finding $\bm{\lambda}^*$ using the stochastic optimization scheme
\begin{align}\label{eq:discrete_time_system}
\bm{\lambda}_{t+1} = \bm{\lambda}_t + \eta \gamma_t \hat{\bm{g}}_t\,,
\end{align}
where $\hat{\bm{g}}_t$ is an unbiased, stochastic estimator of the gradient $\mathcal{L}$ at $\bm{\lambda}_t$
(i.e., $\mathbb{E}\left[\hat{\bm{g}}_t\right] = \nabla \mathcal{L}\left(\bm{\lambda_t}\right)$),
$\eta$ is a base step size, and $\gamma_t > 0$ is the learning rate at iteration $t$, which may depend 
on current and past iterates and gradients. 
The noise in the gradients is a consequence of using mini-batching, 
or approximating the local expectations $\mathcal{L}_i(\bm{\lambda})$ using Monte Carlo estimators, 
or both \cite{mohamed2019monte,hoffman13,ranganath14}. 
For standard stochastic gradient descent (SGD), $\gamma_t$ is a deterministic function of $t$ only and converges asymptotically 
if $\gamma_t$ satisfies the Robbins--Monro conditions 
$\sum_{t=1}^{\infty} \gamma_t = \infty$ and $ \sum_{t=1}^{\infty} \gamma_t^2 < \infty$~\cite{robbins51}. 
SGD is very sensitive to the choice of step size since too large of a step size will result in the algorithm diverging while 
too small of a step size will lead to very slow convergence. 
The shortcomings of SGD have led to the development of more robust, adaptive stochastic optimization schemes such as 
Adagrad~\cite{duchi11}, Adam~\cite{vaswani17, Kingma2014AdamAM}, and RMSProp~\cite{hinton12}, which modify the step size 
schedule according to the norm of current and past gradient estimates.

Even when using adaptive stochastic optimization schemes, however, it remains non-trivial to check for convergence because
we only have access to unbiased estimates of the value and gradient of the optimization objective $\mathcal{L}$. 
Practitioners often run the optimization for a pre-defined number of iterations or use simple moving window statistics 
of $\mathcal{L}$ such as the running median or the running mean to test for convergence. 
We refer to the approach based on looking at the change in $\mathcal{L}$ as the $\Delta$ELBO stopping rule. 
This stopping rule can be problematic as the scale of the ELBO makes it non-trivial to specify a universal convergence tolerance $\epsilon$.
For example, \citet{Kucukelbir15} used $\epsilon = 10^{-2}$, but \citet{yao18a} demonstrate that $\epsilon < 10^{-4}$ might be needed for good accuracy.
More generally, sometimes the objective estimates are too noisy relative to the chosen step size $\eta$, learning rate $\gamma_t$,
threshold $\epsilon$, and the scale of $\mathcal{L}$,
which results in the stopping rule never triggering because the step size is too large relative to the threshold.
The stopping rule can also trigger too early if $\epsilon$ is too large relative to $\eta$ and the scale of $\mathcal{L}$. 
In either case, the user might have to adjust any or all of $\eta$, $\gamma_t$, and $\epsilon$; run the optimiser again;
and then hope for the best.

\subsection{Refining a Variational Approximation}

Another challenge with variational inference is assessing how close the variational approximation $q_{\bm{\lambda}}(\bm{\theta})$ 
is to the true posterior distribution $p$. 
Recently, the $\hat{k}$ diagnostic has been suggested as a diagnostic for variational approximations \citep{yao18a}. 
Let $\bm{\theta}_1,...,\bm{\theta}_S \sim q_{\bm{\lambda}}$ denote draws from the variational posterior. 
Using (self-normalized) importance sampling we can then estimate an expectation under the true posterior as 
$\mathbb{E}\left[f(\bm{\theta)})\right] \approx {\sum_{s=1}^S f(\bm{\theta_s}) w(\bm{\theta}_s)}/{\sum_{s=1}^S w(\bm{\theta}_s)}$,
where $w(\bm{\theta}_s) \equiv {p(\bm{\theta}_s|y)}/{q(\bm{\theta}_s)}$. 
If the proposal distribution is far from the true posterior, the weights $w(\bm{\theta}_s)$ will have high or infinite variance. 
The number of finite moments of a distribution can be estimated using the shape parameter $k$ in the generalized Pareto distribution (GPD) \cite{vehtari2019pareto}. 
If $k>0.5$, then variance of the importance sampling estimate of $\mathbb{E}\left[f(\bm{\theta})\right]$ is infinite. 
Theoretical and empirical results show that values below 0.7 indicate that the approximation is close enough to be used for importance sampling, 
while values above 1 indicate that the approximation is very poor \cite{vehtari2019pareto}. 

Furthermore, recent work \citep{gurbuzbalaban2020heavytail} suggests that SGD iterates converge towards a heavy tailed stationary distribution with infinite variance for even simple models (i.e. linear regression). 
Furthermore, even in cases that don't show infinite variance, the heavy tailed distribution may not be consistent for the mean, i.e. the mean of the stationary distribution might not coincide with the mode of the objective. 
This would imply that some of the assumptions in \citet{Mandt17} wouldn't hold in practice for some parameters in the optimization, which in turn would make iterate averaging unreliable. 
In this work we rely on $\hat{k}$ to provide an estimate of the tail index of the iterates (at convergence) and warn the user when the empirical tail index indicates a very poor approximation. We leave a more thorough study of this phenomenon for future work.

\section{Stochastic Optimization as a Markov Chain}



\Cref{fig:intro-example} (left) shows that as the dimensionality of the variational parameter increases, 
the quality of the variational approximation degrades. 
To understand the source of the problem, we can view a stochastic optimization procedure as producing 
a discrete-time stochastic process $(\bm{\lambda}_t)_{t \geq 1}$ \citep[see, e.g.,][]{Dieuleveut:2020:SGD-MC}. 
Under Robbins--Monro-type conditions, many stochastic optimization procedures converge asymptotically to the exact solution $\bm{\lambda}^*$ \cite{robbins51, li19},  but any iterate $\bm{\lambda}_t$ obtained after a finite number of iterations will be a realization of a diffuse probability distribution $\pi_t$ (i.e., $\bm{\lambda}_t \sim \pi_t(\bm{\lambda}_t)$) 
that depends on the objective function, the optimization scheme, and the number of iterations $t$.

We can gain further insight into the behavior of $(\bm{\lambda}_t)_{t \geq 1}$ by considering an idealized setting -- although see \citet{Dieuleveut:2020:SGD-MC} for a recent more general treatment.
Specifically, \citet{Mandt17} showed that for SGD with constant learning rate (that is, with $\gamma_t = 1$), 
and under simplifying assumptions, the discrete-time stochastic process  $(\bm{\lambda}_t)_{t \geq 1}$ 
defined by \cref{eq:discrete_time_system} can be approximated with a multivariate continuous-time Ornstein--Uhlenbeck process.
The Ornstein--Uhlenbeck process is mean-reverting and admits  stationary distribution, which in 
our setting is
$\pi_\text{OU}(\bm{\lambda}) \equiv \mathcal{N}(\bm{\lambda}|\bm{\lambda}^*, \bm{\Sigma})$~\cite{Mandt17,Vatiwutipong2019}.
%
%
Thus, for some sufficiently large $t_0$, once $t \ge t_0$ the Ornstein--Uhlenbeck process will approximately reach its stationary distribution
and therefore the distribution of  $\bm{\lambda}_t$ satisfies $\pi_t \approx \pi_\text{OU}$. 
This implies that $\bm{\lambda}_t$ is an unbiased estimator of $\bm{\lambda}^*$ with variance $\mathbb{V}[\bm{\lambda}_t] = \bm{\Sigma}$. 

Generally, as the number of model parameters increase -- and hence the number of variational parameters $K$ increases --
the expected squared distance from $\bm{\lambda}$ to the optimal parameter $\bm{\lambda}^*$ will increase. 
For example, assuming for simplicity that the stationary distribution is isotropic with $\bm{\Sigma} = \alpha^2 \bm{I}_K$
(where $\bm{I}_K$ denotes the $K \times K$ identity matrix),
the expected squared distance from $\bm{\lambda}$ to the optimal value is given by
$\mathbb{E}[ \|\bm{\lambda} - \bm{\lambda}^*\|^2] = \alpha^2 K$.
Therefore, we should expect distance between $\bm{\lambda}_t$ and $\bm{\lambda}^*$ 
to be of order $\sqrt{K}$, which implies that the variational parameter estimates output by SGD 
become increasingly inaccurate as the dimensionality of the variational parameter increases. 
As demonstrated in \cref{fig:intro-example}(left), 
one should be particularly careful when fitting a full-rank variational family since the number of parameters is 
$K = P(P+1)/2$. 

Although the preceding discussion only applies directly to SGD, 
it is reasonable to expect that robust stochastic optimization schemes such as Adagrad, Adam, and RMSprop 
will have similar behavior as long as $\gamma_t$ and $\hat{\bm{g}}_t$ depend at most very weakly on iterates far in the past. 
In addition, the theory of \citet{Mandt17} makes additional simplifying that we do not expect to hold exactly in practice.
However, as long as the iterates $(\bm{\lambda}_t)_{t\geq t_0}$ are approximately unbiased and have uniformly bounded variance, 
we expect the idealized theory to be a reasonably good guide to practice. 


\subsection{Improving Optimization Accuracy with Iterate Averaging} \label{sec:iterates_averaging}

While we have shown that we should not expect a single iteration $\bm{\lambda}_t$ to be close to $\bm{\lambda}^*$ in
high-dimensional settings, the expected value of $\bm{\lambda}_t$ \emph{is} equal to (or, more realistically, close to)  $\bm{\lambda}^*$.
Therefore, we can use \emph{iterate averaging} (IA) to construct a more accurate estimate of $\bm{\lambda}^*$ given by 
\begin{align}
\bar{\bm{\lambda}} \equiv \textstyle \frac{1}{T}\sum_{i=1}^T \bm{\lambda}_{t + i},
\label{eqn:iters-rpa}
\end{align}
where we should aim to choose $t \geq t_0$. 
In the idealized setting of \citet{Mandt17}, this estimator is unbiased (that is, $\mathbb{E}[\bar{\bm{\lambda}}] = \bm{\lambda}^*$)
and the variance of $\bar{\bm{\lambda}}$ is 
$\mathbb{V}[\bar{\bm{\lambda}}] = \bm{\Sigma}/T + 2\sum_{1 \leq i < j \leq T}  \text{cov}[\bm{\lambda}_{t + i}, \bm{\lambda}_{t + j}]/T^2$.
Hence, as long as the iterates $\bm{\lambda}_{t}$ are not too strongly correlated, we can reduce the variance and alleviate the effect of dimensionality by using iterative averaging.

Iterate averaging has been previously considered in a number of scenarios. 
\citet{ruppert88} proposes to use a moving average of SGD iterates to improve SGD algorithms in the context of linear one-dimensional models. \citet{polyak92} extend the moving average approach to multidimensional and nonlinear models, and showed that it improved the rate of convergence in several important scenarios; thus, it is often referred to as Polyak--Ruppert averaging. 
In related work, \citet{bach13a} show that an averaged stochastic gradient scheme with constant step size can achieve optimal convergence for linear models even for (non-strongly) convex optimization objectives. 
Recent work demonstrates that averaging iterates can help improve generalization in deep neural networks  \cite{huang17,izmailov18,garipov18,maddox2019simple};
note, however, that our application of IA aims  not just to improve predictive accuracy but also the accuracy of the posterior approximation.

\subsection{Making Iterate Averaging Robust}

In order to make iterate averaging robust in practice, we must (1) ensure that the distributions of the
iterates have finite variance, and (2) determine effective, automatic ways to 
set the two (implicit) free parameters of $\bar{\bm{\lambda}}$: $t$ (when to start averaging) and $T$ (how many iterates to average). 
\#1 is crucial since otherwise even computing a Monte Carlo estimate $\bar{\bm{\lambda}}$ is questionable. 
We use an approach based on the $\hat{k}$ statistic (see \cref{line:check-k-hat-of-iterates} of \cref{algo:workflow}); 
since in our experiments we did not find any cases of infinite-variance iterates, 
we defer further discussion of our approach to the Supplementary Material. 
This use of $\hat{k}$ over the process' iterates is not to be confused with our application of $\hat{k}$ to determine the quality of the variational approximation that we compute after the optimization.
For \#2, recall that our Markov chain perspective suggests that we should start averaging at $t > t_0$,
where $t_0$ denotes the iteration after which the distribution of $\bm{\lambda}_t$ has approximately reached stationarity
and therefore is near the optimum \citep{roux19,jain18}. 
We must then select $T$ large enough that  $\bar{\bm{\lambda}}$ is sufficiently close to $\bm{\lambda}^*$.
We address how to robustly choose $t$ and $T$ in turn. 

\paragraph{Determining when to start averaging}
Previous approaches to selecting $t$ rely on the so-called Pflug criterion \cite{Pflug1990, roux12,chee18a},
which is based on evaluating the sum of the inner product of successive gradients. 
Unfortunately this approach is not robust and can be slow to detect convergence \citep{Pesme:2020}. 
To develop an alternative, robust approach to selecting $t$ we turned to the Markov chain Monte Carlo literature.
In MCMC, the $\widehat{R}$ statistic is a canonical way to determine if a Markov chain have reached stationarity~\cite{gelman92,BDA3,vehtari2019rhat}. 
The standard approaches to computing $\widehat{R}$ is to use multiple Markov chains.
If we have $J$ chains and $N$ iterates in each chain, $\bm{\lambda}^{(j)}_{i}$, such that $i = 1, \ldots, N; j = 1, \ldots, J$, 
then
$\widehat{R} \equiv  (\hat{\mathbb{V}}/\hat{\mathbb{W}})^{1/2}$,
where $\hat{\mathbb{V}}$ and  $\hat{\mathbb{W}}$ are estimates of, respectively, 
the between-chain and within-chain variances.
We use the split-$\widehat{R}$ version, where all chains are split into two before carrying out the computation above, which helps with detecting non-stationarity \citep{BDA3,vehtari2019rhat} 
and allows us to use it even when $J = 1$. 

In order to utilize $\widehat{R}$, we run $J$ optimization runs (``chains'') in parallel and consider
the iterates at stationarity when $\widehat{R} < \tau$, where $\tau > 1$ is a user-chosen cutoff. 
We select a moving window $W$ and only use the most recent $W$ samples for computing $\widehat{R}$
since we do not expect iterates before the (unknown) $t_0$ to be close to the stationary distribution. 
There is a trade-off between making $W$ large, which leads to more accurate and potentially smaller 
estimates for $\widehat{R}$, and making $W$ small, which leads to more quickly determining when the iterates are near stationarity. 
In practice we found $W = 100$ to be a good choice, although somewhat larger or smaller values
would work as well. 
Concerning the choice of the cutoff $\tau$, 
in the MCMC literature $\widehat{R}$ is required to be very precise since the stationary distribution is the true posterior,
so $\tau = 1.01$ or even smaller is recommended \citep{vehtari2019rhat,Vats:2018:Rhat}.
In our case, because we use a fairly small value for $W$ and are less concerned 
about the quality of the stationary distribution, we use $\tau = 1.1$.
 



%

\paragraph{Determining when to stop averaging}

Once $t > t_0$ is found using $\widehat{R}$, we must determine how many iterates to average.
Since all $J$ optimizations are guaranteed 
to reach the same optimum (if there are no local optima) due to our use of $\widehat{R}$,
we can combine the iterates into a single variational parameter estimate 
$\bar{\bm{\lambda}} = \sum_{j=1}^J\sum_{i=1}^T \bm{\lambda}^{(j)}_{t + i} / (JT)$,
where $\bm{\lambda}^{(j)}_{s}$ the $s$th iterate of the $j$th chain. 

Due to the non-robustness of the  $\Delta$ELBO stopping rule, 
we propose an alternative stopping criterion that is robust to the (unknown) scale of the objective
and which accounts for the fact that the variational parameter is the quantity of interest, not the value of
the objective function.
Again turning to the MCMC literature and taking advantage of our iterative averaging approach,
we propose to use the Monte Carlo standard error (MCSE)~\cite{Hastings:1970,flegal08b,vehtari2019rhat},
which is given as
   $\textrm{MCSE}(\lambda_i) \equiv \{\mathbb{V}(\lambda_i)/{\textrm{ESS}(\lambda_i)}\}^{1/2}$, 
where 
$\mathbb{V}(\lambda_i)$ is the variance of the $i$th component of the iterates,
 $\textrm{ESS} \equiv {JN}/(1 + \sum_{t=1}^{\infty}2\rho_t)$ is the effective sample size (ESS),
$N$ is the number of iterations after $\widehat{R}$ convergence (used to compute the variance), 
and $\rho_t$ is the autocorrelation at lag $t$. 
The ESS accounts for the dependency between iterates and in general we expect it to be smaller than the 
total number of iterates $JN$.
We compute the ESS using the method described in \citet{vehtari2019rhat}. 
In addition to checking that the median value of the $\textrm{MCSE}(\lambda_i)$ 
is below some tolerance $\epsilon$, 
to ensure the MCSE estimates are actually reliable, we also require that all of the 
effective sample sizes are above a threshold $e$.

We note that a benefit of our approach is that the MCSE also provides an estimate of how many significant figures in the 
parameter estimate $\bar{\bm{\lambda}}$ are reliable.
Such reliability estimates are particularly important in high dimensions since, as we will see (\cref{sec:experiments} and \cref{tbl:results-ds1}),
even small perturbations to the location or scale parameters can result in a very bad approximation 
to the posterior distribution.

\paragraph{Diagnosing convergence problems with autocorrelation values}
The autocorrelation values $\rho_t$ that are computed when estimating ESS can also used as a diagnostic 
if $\widehat{R}$ is not falling below $\tau$ or the MCSE is not decreasing when more iterations are averaged. 
Large autocorrelations before $\widehat{R} < \tau$ may indicate that the window $W$ needs to be increased 
in order to estimate $\widehat{R}$ effectively.
Large autocorrelations after averaging has started suggests iterate averaging may not be reliable. 
\begin{algorithm}[t]
\caption{Robust Stochastic Optimization for Variational Inference}
\label{algo:workflow}
\begin{algorithmic}[1]
\State\textbf{Input:} learning rate $\eta$, \# of optimization runs $J$, window size $W$, $\widehat{R}$ cutoff $\tau$, 
MCSE cutoff $\epsilon$, ESS cutoff $e$, iterate initalizations $\bm{\lambda}_0^{(j)}$ for $j=1,\dots,J$
\For{$t \gets 1$ to $T_{\mathrm{max}}$}
    \State Compute $\bm{\lambda}_t^{(j)}$ via \cref{eq:discrete_time_system}, $j=1,\dots,J$
    \If {$t \bmod W = 0$}
        \State Compute $\widehat{R}_i$, the $\widehat{R}$ value for the $i$th component of $\bm{\lambda}$ 
        \If{$\max_i \widehat{R}_i < \tau$}
            \State $T_{0} \gets t$
            \State \textbf{break}
        \EndIf
    \EndIf
\EndFor
\If{$\max_i \widehat{R}_i < \tau$ or $\hat{k}$ of iterates $> 1.0$} \label{line:check-k-hat-of-iterates}
    \State Warn user that optimization may not have converged 
    \State \Return $\bar{\bm{\lambda}}$ computed from the last $W$ iterates 
\Else 
        \For{$t \gets T_{0}$ to $T_{\textrm{max}}$} 
        \State Compute $\bm{\lambda}_t^{(j)}$ via \cref{eq:discrete_time_system}, $j=1,\dots,J$
         \If {$t - T_0 \bmod W = 0$ and $\textrm{MCSE} < \epsilon$ and $\textrm{ESS} > e$} \Comment{using last $t - T_0$ iterates}
         \State \textbf{break}
         \EndIf
         \EndFor
        \State \Return $\bar{\bm{\lambda}}$ computed from the last $t - T_0$ iterates
    \EndIf
\end{algorithmic}
\end{algorithm}


\begin{figure}[t]
\begin{center}
\begin{subfigure}[t]{.35\textwidth}
    \includegraphics[width=\textwidth]{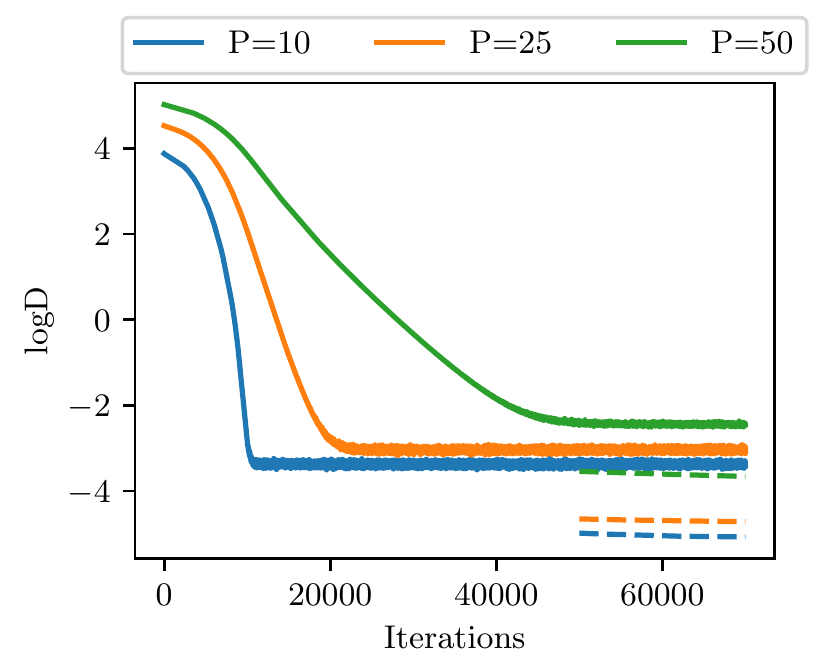}
    \caption{}
    \label{fig:linreg-FR-accuracy}
\end{subfigure}
\hfill
\begin{subfigure}[t]{.36\textwidth}
    \includegraphics[width=\textwidth]{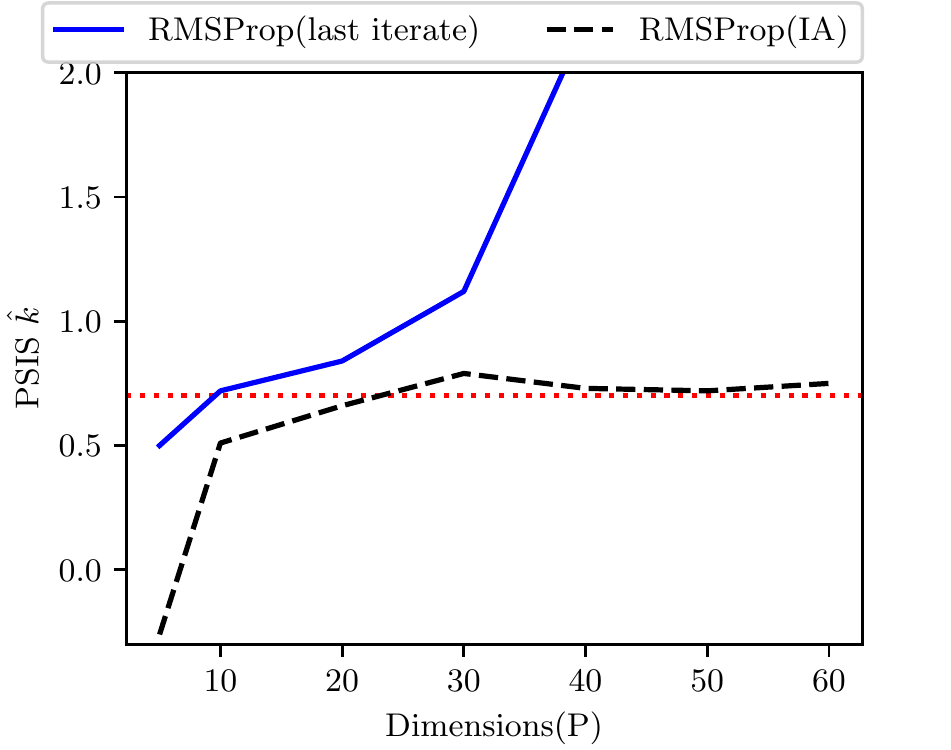} 
    \caption{}
    \label{fig:linreg-FR-k-hat}
\end{subfigure}
\hfill
\begin{subfigure}[t]{.26\textwidth}
    \includegraphics[width=\textwidth]{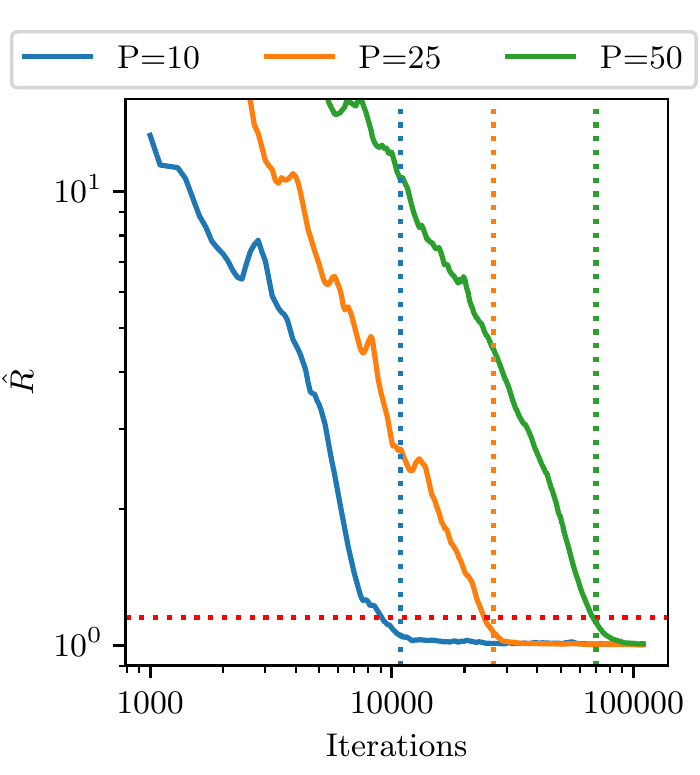}
    \caption{}
    \label{fig:linreg-FR-R-hat}
\end{subfigure}
\end{center}
\caption{
For the linear regression model with posterior correlation $0.9$, 
the evolution of \textbf{(a)}  moment distance $D$, \textbf{(b)} $\hat k$ statistic, and \textbf{(c)} $\widehat{R}$ statistic
during optimization.
For $D$ and $\hat{k}$ (of the variational approximation) we show the values for the last iterate (solid lines) and averaged iterates (dashed lines).}
\label{fig:linreg-FR}
\end{figure}

\section{Experiments}\label{sec:experiments}

We now turn to validating our robust stochastic optimization algorithm for variational inference
(summarized in \cref{algo:workflow}) 
through experiments on both simulated and real-world data.
In our experiments we used $\eta = 0.01, W = 100, \tau = 1.1$, and $e=20$.
To ensure a fair comparison to the $\Delta$ELBO stopping rule, we used $J = 1$ in all of our experiments;
the exception is that \cref{fig:linreg-FR} used $J=4$ since it does not involve a comparison to $\Delta$ELBO. 
We also put $\Delta$ELBO at an advantage by doing some tuning of the threshold $\epsilon$, while keeping 
$\epsilon = 0.02$ when using our MCSE criterion. 
We show the results based on using RMSprop, but we found that AdaGrad performed similarly (see Supplementary Material).
For the variational approximation family we used multivariate Gaussians $q(\bm{\theta}) = \mathcal{N}(\bm{\theta}; \bm{m}_{q}=\bm{\mu}, \bm{\Sigma}_{q}=\bm{LL}^T)$ where $\bm{L}$ is the Cholesky decomposition of the covariance matrix. 
We used \texttt{viabel}~\citep{Huggins:2020:VI} for inference, TensorFlow Probability~\citep{tfp} and Stan~\citep{Carpenter:2017} for model-construction, and \texttt{arviz}~\citep{arviz2019} for tail-index estimation. 

\begin{table*}[b]
\footnotesize
\setlength {\tabcolsep}{3pt}
\centering
\caption{
Real-data results comparing the $\Delta$ELBO stopping rule to our proposed MCSE stopping rule (which implements all of \cref{algo:workflow}).
$K =$ number of variational parameters, $\epsilon =$ threshold used for the stopping rule,
and $T =$ total number of iterations before termination. 
* denotes that convergence was not reached after $T_{\max}$ iterations.
}
\begin{tabular}{l|c|cc|c|cc|cc|cc}
 \textbf{Model} & K  & Stopping rule & $\epsilon$ &  $T$ & $D_{\bm{\mu}}$  & $D_{\bm{\mu}}$ (IA)  & $D_{\bm{\Sigma}}$  &  $D_{\bm{\Sigma}}$ (IA) & $\hat{k}$ & $\hat{k}$(IA)  \\
 \hline
  Boston
 & 104 & $\Delta$ELBO & 0.01 & 2100 & 0.02 & 0.008 & 0.06 & 0.38 & 0.90 & 11.2   \\
  &  & MCSE & 0.02 & 5900 & 0.003 & $\textbf{0.001}$ & 0.008  & $\textbf{0.004}$ & 0.55 & $\textbf{0.06}$ \\
 \hline
   Wine 
  & 77 & $\Delta$ELBO & 0.001 & 2400 & 0.005 & 0.004 & 0.017 & 0.11 & 0.78 & 15     \\
  & & MCSE & 0.02 & 5300 & 0.002 & \textbf{0.001} & 6e-5 & $\textbf{3e-5}$ & 0.70 & \textbf{0.07} \\
 \hline
Concrete 
  & 44 & $\Delta$ELBO & 0.005 &  1800 & 0.02  & 0.04 & 0.018 & 0.51 & 2.65 & 15.4    \\
 &  & MCSE & 0.02 &  3900 & 0.015  & \textbf{0.001} & 0.02 & \textbf{0.004} & 0.74 & \textbf{0.09}    \\
 \hline
8-school (CP) & 65 & $\Delta$ELBO & 0.005 & 1100 & 1.9 & 4.5 & \textbf{3.5} & 5.8 & 0.98 & 0.85    \\
 & & MCSE & 0.02 & 6200 & \textbf{1.8}  & 2.1 & \textbf{3.5} & 3.7 & 0.88 & \textbf{0.78}    \\
 \hline
8-school (NCP) 
 & 65 & $\Delta$ELBO & 0.005 & 1700 & 0.12  & \textbf{0.09} & 1.02 & 1.02 & 0.60 & 0.60    \\
 & & MCSE & 0.02 & 2400 & 0.14  & 0.13 & 1.05 & \textbf{0.98} & \textbf{0.58} & 0.63    \\
 \hline
Radon 
& 4094 & $\Delta$ELBO & 0.02 & 15000* &  5.8 & \textbf{5.7} & 0.80 & \textbf{0.40} & 1.2 & \textbf{0.34}    \\
 &  & MCSE & 0.01 & 9500 & 6.0 & 5.9 & 1.2 & 1.1 & 1.3 & 0.40  \\
 \hline
MNIST100 
  & 7951 & $\Delta$ELBO & 0.001 & 1200 & 82.7  & 83.7 & 34.1 & 34.1 & 32.59 & 31.93 \\
  & & MCSE & 0.02 & 10000* & {\bf 33.6} & 51.0  & \textbf{34.0} & \textbf{34.0} & {\bf 6.94} & 10.57 \\
\end{tabular}
\vspace{4pt}
\label{tbl:results-ds1}
\end{table*}

The linear regression experiments with synthetic data mentioned in \cref{sec:intro} (and described in detail in the Supplementary Material)
provide a useful case study of stochastic variational inference where the true posterior distribution
belongs to the variational family, meaning that any inaccuracy in the variational approximation was due to the
stochastic optimization procedure.
We also investigated a variety of models and datasets using black box variational inference:
logistic regression~\citep{yeh1998modeling} on three UCI datasets (Boston, Wine, and Concrete~\cite{Dua2019});
a high-dimensional hierarchical Gaussian model (Radon \cite{lin99}), 
the 8-school hierarchical model~\cite{rubin81},
and a Bayesian neural network model with 10 hidden units and 2 layers \citep{lampinen2001bayesian} 
to classify 100 handwritten digits from the MNIST dataset~\citep{lecun1998gradient} (MNIST100).
The 8-school model has a significantly non-Gaussian posterior
and has served as a test case in a number of recent variational inference papers \citep{yao18a,Huggins:2020:VI}.
We considered both the centered parameterization (CP) and non-centered one (NCP)
because the NCP version of 8-school is easier to approximate with variational methods \citep{yao18a,Huggins:2020:VI},
and therefore experiments on both provide insight into the robustness of a variational algorithm. 
For all real-data experiments we estimated the ground-truth posterior moments 
(i.e., the mean $\mu$ and covariance matrix $\Sigma$) 
using the dynamic Hamiltonian Monte Carlo algorithm in Stan \citep{Carpenter:2017}.
We used these to compute the normalized moment distance $D \equiv (D_{\bm{\mu}}^2 + D_{\bm{\Sigma}}^2)^{1/2}$, where $D_{\bm{\mu}} \equiv \|\bm{\mu} - \hat{\bm{\mu}}\|_2$,
$D_{\bm{\Sigma}} \equiv \|\bm{\Sigma} - \hat{\bm{\Sigma}}\|^{1/2}$  
and $\hat{\bm{\mu}}$ and $\hat{\bm{\Sigma}}$ denote, respectively,
the variational estimates of the posterior mean and covariance.

\paragraph{Iterate averaging improves variational parameter estimates}
First we investigated the benefits of using iterate averaging rather than the final iterate.
For the linear regression model,
\cref{fig:intro-example} shows the benefits of IA when using either $\Delta$ELBO or MCSE
as a stopping criteria, with a larger gain coming from its use 
with MCSE (and $\widehat{R}$) since in that case the iterates were closer to the optimum. 
\Cref{fig:intro-example}(right) shows the improved accuracy of iterate averaging compared 
to using the last iterate in detail for the case when the dimension of the linear regression model was $P=70$. 
\Cref{fig:linreg-FR-accuracy,fig:linreg-FR-k-hat} provides a further example of the benefits of 
iterate averaging for linear regression in the more challenging case of strong posterior correlation.
IA provides an approximately two orders of magnitude improvement in accuracy.
The improvement in importance sampling performance is also dramatic:
while the $\hat{k}$ statistic for the variational approximation after the last iterate is above the 0.7 reliability threshold even
when with data of dimension $P = 10$, the $\hat{k}$ statistic of IA remains below or near the $0.7$
when $P = 60$. 

\Cref{tbl:results-ds1} shows that in our real-data experiments, 
IA almost universally outperforms the last iterate when using \cref{algo:workflow}, 
both in terms of moment estimates and approximation's $\hat{k}$; 
however, because the $\Delta$ELBO stopping rule sometimes resulted
in premature termination of the optimizer, IA did not always provide a benefit with $\Delta$ELBO,
which lends further support for using our more comprehensive robust optimization framework. 
The only exception was the (multimodal) MNIST100 posterior, where for MCSE 
the $\hat k$ statistic for the last iterate was superior to that for IA -- although 
both were very large. 

\paragraph{MCSE stopping criteria improves robustness and accuracy}
Recall that \cref{fig:intro-example} (left)  
provides an case where the $\Delta$ELBO stopping rule results in premature termination of the optimizer.
For the real-data examples, in \cref{tbl:results-ds1} we see that due to substantially earlier termination
(small $T$), using $\Delta$ELBO consistently results
is less accurate posterior approximations in terms of moment estimates and $\hat{k}$.
The only exception is the Radon model, which never reaches convergence according to 
the $\Delta$ELBO criterion and, as a result, produces better posterior mean accuracy
and a smaller $\hat{k}$ statistic than using MCSE. 
On the other hand, MCSE runs for approximately a third as many iterations, still
has a $\hat{k}$ statistic less than 0.5, and produces a more accurate posterior mean estimate. The threshold $\epsilon$ was kept the same for all the datasets in case of MCSE, and we found it to be quite robust in contrast to

\paragraph{Autocorrelation and $\hat{k}$ detect problematic variational approximations}
Supplementary Figure 1a provides an example where, for linear regression, the oscillation in 
the autocorrelation plot indicates super-efficiency in the averaging due to negative correlation in odd lags \citep{vehtari2019rhat}.
Supplementary Figures 1b and 1c
provide examples where, for the 8-school models (both CP and NCP), the iterates are heavily correlated
and thus averaging is less efficient,
which is reflected in the less dramatic benefits of using IA (\cref{tbl:results-ds1}). 
The $\hat{k}$ statistics (\cref{tbl:results-ds1}) provide good guidance
of approximation accuracy. 


\paragraph{$\widehat{R}$ detects optimization failure}
\Cref{fig:intro-example,fig:linreg-FR-R-hat} and \cref{tbl:results-ds1} provide examples
where $\widehat{R}$ successfully detects convergence of the optimization.
 Just as importantly, $\widehat{R}$ can also diagnose optimization problems such
as multi-modality. 
For example, if the variational objective has multiple (local) optima, different 
optimizations can end up in different optima due to by random initialization; but
this would be indicated by a large $\widehat{R}$.  
For example, when we used \cref{algo:workflow} with $J=4$ for the multimodal MNIST100 model, the maximum $\widehat{R}$ 
was $4.8$. 
This result also provides support for using $J > 1$ parallel optimizations, since such multimodality 
cannot be detected when $J = 1$. 
A direction for future work would be to approximate a multimodal posterior by extending our 
approach to analyze the convergence in each mode and then combine results of different modes 
(e.g., by stacking weights~\citep{yao18a}).
\section{Acknowledgements}
We would like to thank Ben Bales for useful discussions about $\widehat{R}$ statistic.

\bibliography{iwvi,jonathan}
\bibliographystyle{eabbrvnat}
\setcounter{section}{1}

\renewcommand{\thesubsection}{\Alph{section}\arabic{subsection} }

\newpage
\onecolumn
\section*{Appendix}

\subsection{Monte Carlo Gradients in Stochastic Optimization}

The exact gradient of the ELBO is given by
\begin{align}
\nabla \mathcal{L}\left(\bm{\lambda}_t\right) = \sum_{i=1}^N \nabla \mathcal{L}_i\left(\bm{\lambda}\right).  
\end{align}
There are two possible sources of stochasticity in the gradient estimation: 1) use of mini-batches of data and 2) Monte Carlo estimates of ELBO as in black box variational inference (BBVI).
The mini-batch approximation is given by
\begin{align}\label{eq:gradient_minibatch}
    \hat{\bm{g}}^{\text{MB}}_t = \frac{N}{|\mathcal{S}|}\sum_{s \in \mathcal{S}}  \nabla \mathcal{L}_s\left(\bm{\lambda}_t\right),
\end{align}
where $\mathcal{S}$ is an index set for a random subset of the observations. In BBVI, the local expectations $\mathbb{E}_{q} \left[\ln p(y_i| \bm{\theta})\right] \approx \frac{1}{M}\sum_{m=1}^M \ln p(y_i|\bm{\theta}_m)$ are estimated using $M$ Monte Carlo draws $\bm{\theta}_m \sim q_{\bm{\lambda}}$ as
\begin{align}\label{eq:gradient_mc}
\hat{\bm{g}}^{\text{MC}}_t &\approx \frac{1}{M}\sum_{m=1}^M\sum_{i=1}^N \left(\nabla\ln p(y_i| \bm{\theta}_m) - \frac{1}{N}\nabla\ln \frac{q(\bm{\theta}_m)}{p_0(\bm{\theta}_m)}\right).
\end{align}

\subsection{Further Details for Section 3}

Recall in our discussion of the implications of \citet{Mandt17}, we assumed for simplicity that the stationary distribution 
of SGD is isotropic; that is, that $\bm{\Sigma} = \alpha^2 \bm{I}$.
It follows that the squared distance from $\bm{\lambda}$ to the optimal value $\bm{\lambda}^*$ is given by
\begin{align}
\label{eq:A}
A = \|\bm{\lambda} - \bm{\lambda}^*\|^2 = \alpha^2 \|\bm{z}\|^2 = \alpha^2 \bm{z}^T \bm{z} = \alpha^2 \sum_{k=1}^K z_k^2,
\end{align}
where $\bm{z} \sim \mathcal{N}(0, \bm{I})$. 
It follows that the expected squared distance to the mode is  $\mathbb{E}[A] = \alpha^2 K$.
The corresponding expected squared distance for the proposed estimator $\bar{\bm{\lambda}}$ is given by
\begin{align}
\label{eq:barR}
\bar{A} = \|\bar{\bm{\lambda}} - \bm{\lambda}^*\|^2 = \| \frac{1}{T}\sum_{t=1}^T \left(\bm{\lambda}^* + \alpha \bm{z}_t\right)  - \bm{\lambda}^*\|^2,
\end{align}
where $\bm{z}_t \sim \mathcal{N}(\bm{0}, \bm{I})$. It follows that $\mathbb{E}\left[\bar{A}\right] = \alpha^2 K/T$ and thus, using the estimator $\bar{\bm{\lambda}}$ reduces the expected square distance by a factor of $T$ when the iterates are i.i.d. However, in practice, the iterates will correlated and the rate of decrease will be slower. The variance of $\bar{\bm{\lambda}}$ is then given by
\begin{align}
\mathbb{V}\left[\bar{\bm{\lambda}}\right] 
= & \frac{1}{T} \bm{\Sigma} + \frac{2}{T^2}\sum_{1 \leq i < j \leq T}  \text{cov}\left[\bm{\lambda}_{t + i}, \bm{\lambda}_{t + j}\right].
\end{align}

\subsection{Stochastic process tail index diagnostic}
In cases where the assumptions given in Section 3
are not obeyed, we cannot obtain reliable Monte Carlo estimates via iterative averaging:
as given by the central limit theorem, the stationary distribution should have finite variance in order for averaging to work 
(or finite mean for the generalized central limit theorem).
A robust way to detect distributions with heavy tails is the Pareto-$\hat{k}$ diagnostic given in \citet{vehtari2019rhat}.
The $\hat{k}$ diagnostic operates by fitting a generalized Pareto distribution to a single tail of a sample.
Specifically, $\hat{k}$ is the estimated shape parameter $k$, that determines that the distribution has moments up to the $(1/k)$th. 
We compute $\hat{k}$ for the lower and upper tails of each component of $\bm{\lambda}_t$.
\citet{vehtari2019rhat} provide theoretical and experimental justification that small error rates can be achieved in 
averages under the generalized central limit theorem if the tail index $k<0.7$.
Because $\hat{k}$ estimates tend to be conservative and we are often computing a large number of them,
we determined that any $\hat{k}$ value greater than $1$ to be reported as problematic in our experiments. The maximum value of $\hat{k}$ index over all the variational parameters for the linear regression model was found to be 0.12, for the eight school models non-centred parameterization it was found to be 0.09, and with centred parameterization it was found to be 0.40. Since these values were less than the threshold of 1 as reported in the main text, we proceeded with our experiments and use the iterate averaging workflow. Since, this value is related to the gradient variance, the analysis of different models with different divergence measures will form potential future work.





\subsection{Additional Details for Bayesian Linear Regression Experiments}

We now describe the Bayesian linear regression model used in our experiments in detail.
We use a Gaussian prior for the regression coefficients $\beta$ and known noise variance $\sigma^2$ so 
that the posterior is Gaussian. 
Therefore the only error in the approximation is explained by the optimization. We compare the standard optimizer solutions to our proposal in a variety of configurations.

The assumed generative model is $\bm{y} \mid \bm{X} \sim \mathcal{N}(\bm{X}\bm{\beta}, \sigma^2)$ and ${\beta}_k \sim \mathcal{N}(0,1)$ 
with $\sigma^2 = 0.4$ fixed.
We generated data from the same model with covariates for each sample generated according to 
$(x_{nP}, \ldots, x_{nP}) \sim \mathcal{N}(0,K)$, where $K_{ij} = \gamma^{|i-j|}$.
Note that correlation $\gamma$ in the design matrix imposes a  correlation structure in the posterior. 
In our experiments we fix the sample size $N = 300$ and vary the dimension $P$.
%
To account for randomness in the simulations we average the results over 50 data realizations of $\bm{X}$, $\bm{\beta}$, and $\bm{y}$.
We used $T_{\textrm{max}}=120~000$ iterations/20000 epochs  (complete passes over the data) with minibatch size $|\mathcal{S}|=50$ datapoints. 


\subsection{Additional Results}

The results in Fig. 2 are replicated in \cref{fig:linreg-FR-R-hat2} using $\gamma=0.5$ rather than $\gamma = 0.9$.
The results in Table 1 are replicated in \cref{tbl:adagrad-results} using Adagrad rather than RMSprop. 

\begin{figure}[t]
\begin{subfigure}[t]{.38\textwidth}
    \includegraphics[width=\textwidth]{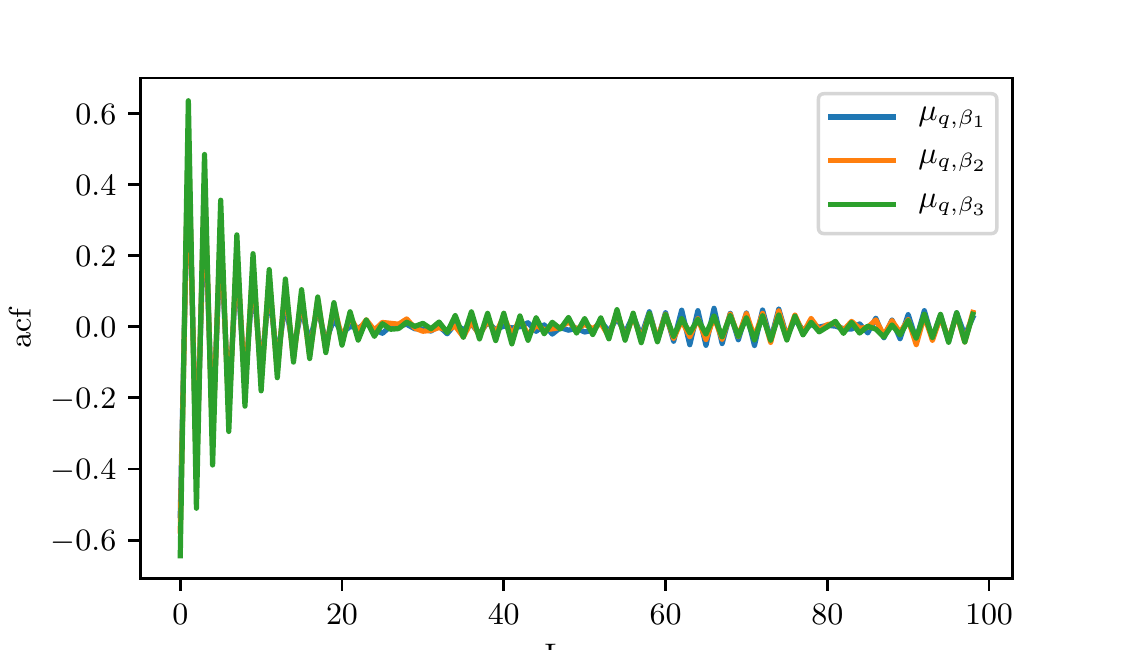}
    \caption{Linear Regression}
    \label{fig:autocorrelation-linreg}
\end{subfigure}
\begin{subfigure}[t]{.28\textwidth}
    \includegraphics[width=\textwidth]{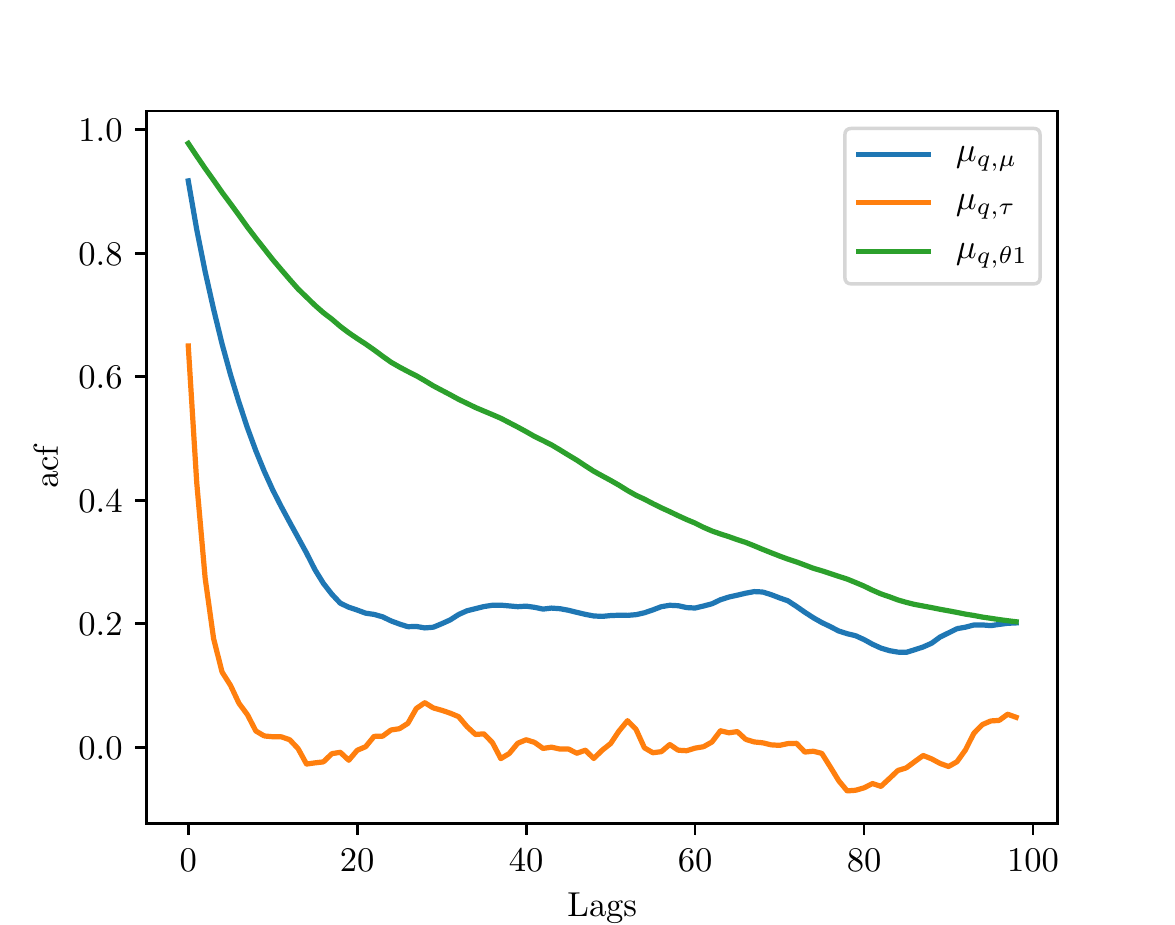}
    \caption{8-school (CP)}
    \label{fig:autocorrelation-8-school-CP}
\end{subfigure}
\begin{subfigure}[t]{.28\textwidth}
    \includegraphics[width=\textwidth]{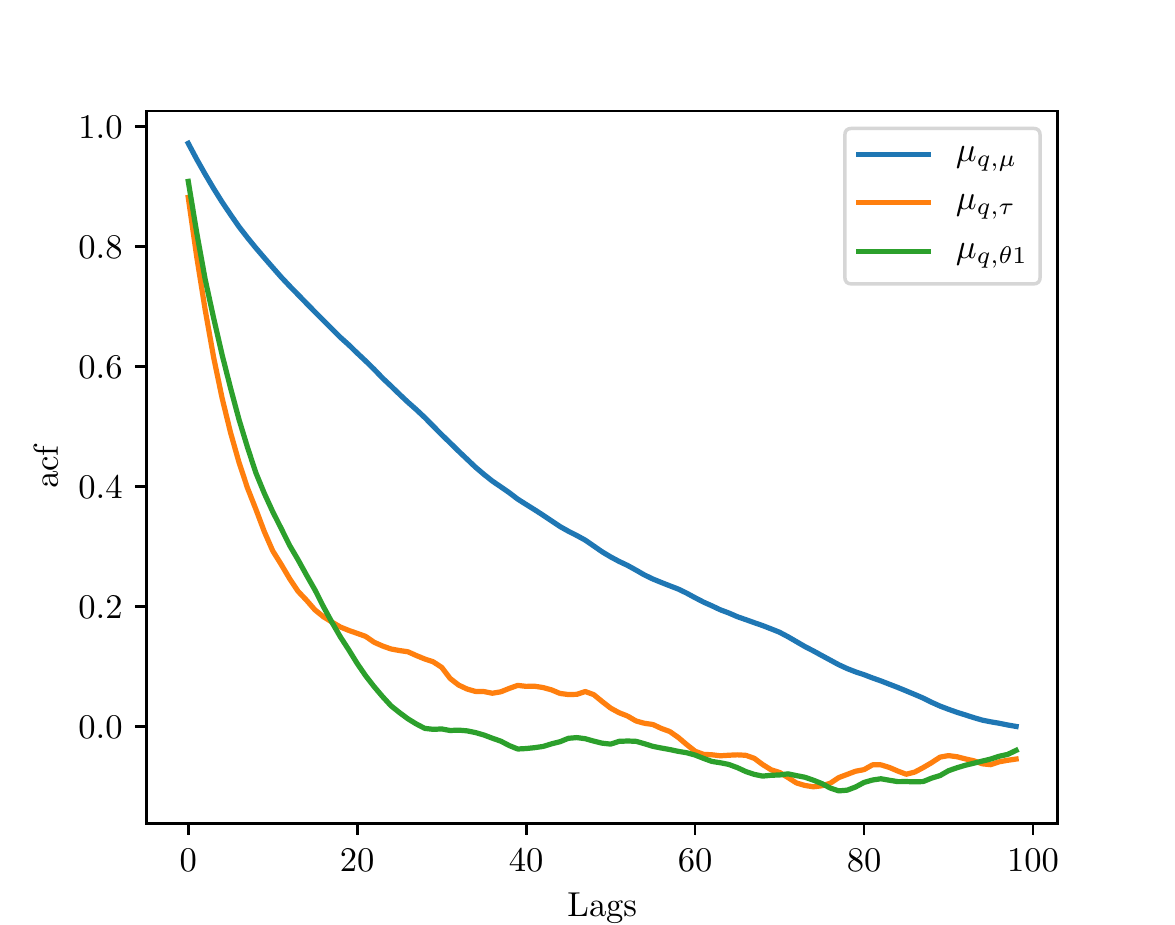}
    \caption{8-school (NCP)}
    \label{fig:autocorrelation-8-school-NCP}
\end{subfigure}
\caption{Autocorrelation plots for
\textbf{(a)} the location parameters for $\beta_1$, $\beta_2$, and $\beta_3$ for
linear regression using a mean-field variational family and
\textbf{(b,c)} the location parameters of $\mu, \tau$ and $\theta_1$ for 
8-schools centered and non-centered parameterisations. The plots serve as a diagnostic tool for assessing the efficiency of averaging. }
\label{fig:autocorrelation}
\end{figure}

\begin{table*}[tp]
\footnotesize
\setlength {\tabcolsep}{5pt}
\centering
\caption{
Comparison of stopping rules on different datasets with different optimisers, where we begin averaging after approximate convergence using Rhat statistic. We used Adagrad to obtain these results.}
\begin{tabular}{l|c|cc|c|cc|cc|cc}
 \textbf{Model} & K & Stopping rule & $\epsilon$ &  $T$ & $D_{\mu}$  & $D_{\mu}$ (IA)  & $D_{\Sigma}$  &  $D_{\Sigma}$ (IA) & $\hat{k}$ & $\hat{k}$(IA)  \\
 \hline
  Boston
 & 104 & $\Delta$ELBO & 0.01 & 1200 & 0.01 & 0.008 & 0.33 & 0.37 & 13.9 & 16.2   \\
  &  & MCSE & 0.02 & 4700 & 0.005 & \textbf{0.002} & 0.02  & \textbf{0.01} & 0.40 & \textbf{0.03}  \\
 \hline
  Wine 
  & 77 & $\Delta$ELBO & 0.002 & 1800 & 0.008 & 0.001 & 0.06 & 0.08 & 1.5 & 1.9     \\
  & & MCSE & 0.02 & 7000 & 0.004 & \textbf{0.001} & 0.013 & \textbf{0.006} & 0.65 & \textbf{0.01} \\
 \hline
Concrete & 44 & $\Delta$ELBO  & 0.02 &  1900 & 0.009  & 0.002 & 0.17 & 0.22 & 3.6 & 4.5    \\
 & & MCSE & 0.02 &  7900 & 0.004  & \textbf{0.001} & 0.008 & \textbf{0.006} & 0.68 & \textbf{0.02}    \\
 \hline
8-school (CP) & 65 & $\Delta$ELBO & 0.01 & 3800 & 11.0  & 11.1 & 7.9 & 7.9 & \textbf{0.95} & 0.99    \\
 & & MCSE & 0.02 & 15000* & \textbf{5.7}  & 7.1 & \textbf{4.2} & 5.5 & \textbf{0.90}  & 0.96    \\
 \hline
8-school (NCP) & 65 & $\Delta$ELBO & 0.01 & 1800 & 2.6  & 2.7 & \textbf{0.90} & 0.91 & 0.65 & 0.60    \\
 & & MCSE & 0.02 & 5600 & 0.09  & \textbf{0.07} & 0.97 & 0.96 & 0.62 & \textbf{0.55}  \\

\end{tabular}
\vspace{4pt}
\label{tbl:adagrad-results}
\end{table*}


\begin{figure}[t]
\begin{center}
\begin{subfigure}[t]{.35\textwidth}
    \includegraphics[width=\textwidth]{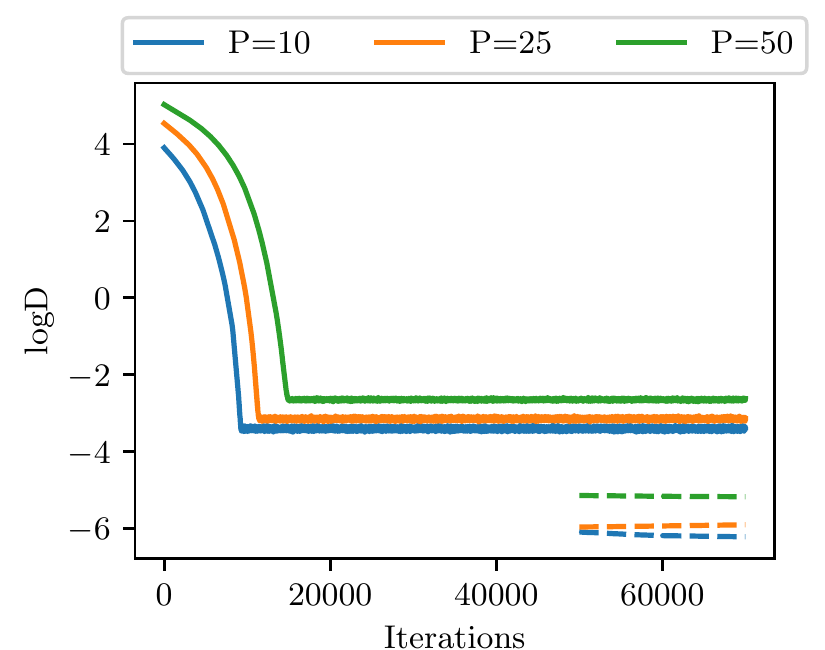}
    \caption{}
    \label{fig:linreg-FR-accuracy}
\end{subfigure}
\hfill
\begin{subfigure}[t]{.36\textwidth}
    \includegraphics[width=\textwidth]{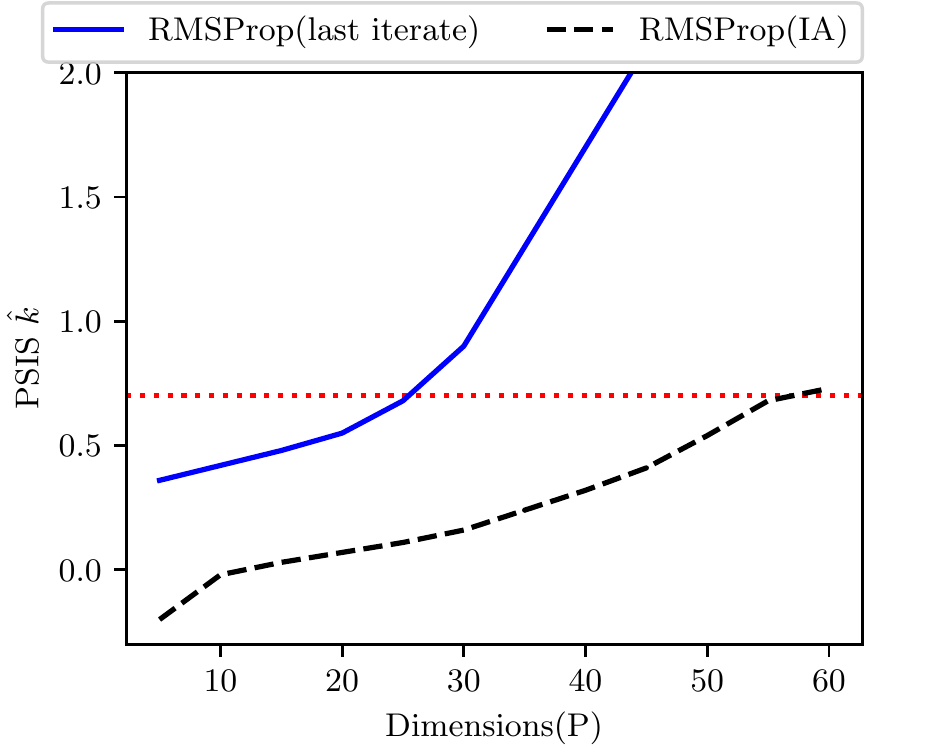} 
    \caption{}
    \label{fig:linreg-FR-k-hat}
\end{subfigure}
\hfill
\begin{subfigure}[t]{.26\textwidth}
    \includegraphics[width=\textwidth]{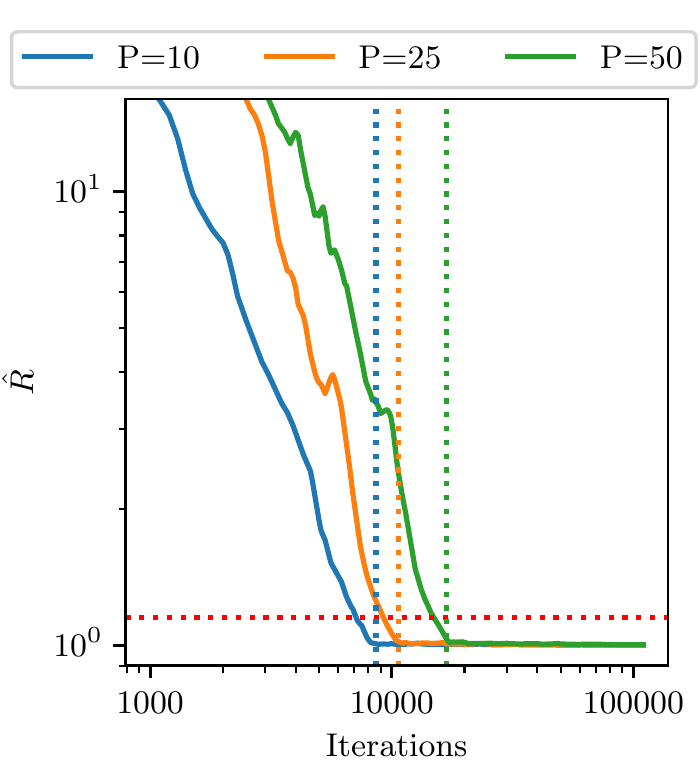}
    \caption{}
    \label{fig:linreg-FR-R-hat2}
\end{subfigure}
\end{center}
\caption{
For the linear regression model with posterior correlation $0.5$, 
the evolution of \textbf{(a)}  moment distance $D$, \textbf{(b)} $\hat k$ statistic, and \textbf{(c)} $\widehat{R}$ statistic
during optimization.
For $D$ and $\hat{k}$ we show the values for the last iterate (solid lines) and averaged iterates (dashed lines). The convergence here happens earlier than with $0.9$ correlation shown in main text, which can be seen from both (a) and (c) plots.}
\label{fig:linreg-FR-all}
\end{figure}

\end{document}